\documentclass{article}
\usepackage[accepted]{icml2025}
\usepackage{amsmath}
\usepackage{natbib}
\usepackage{graphicx}
\usepackage{pdfpages}
\usepackage{float}
\usepackage{amsthm}
\usepackage{amsmath}
\usepackage{amssymb}
\usepackage{mdframed}
\usepackage{xcolor}
\usepackage{cuted}
\usepackage{graphicx}
\usepackage{booktabs}
\usepackage{caption}
\usepackage{subcaption}
\usepackage{xcolor}
\usepackage{hyperref}

\definecolor{froggergreen}{RGB}{0, 128, 0}  

\newmdenv[
  linecolor=froggergreen,
  linewidth=2pt,
  topline=true,
  bottomline=true,
  leftline=true,
  rightline=true,
  backgroundcolor=green!5,  
  innertopmargin=15pt,
  innerbottommargin=15pt,
  innerleftmargin=15pt,
  innerrightmargin=15pt,
  roundcorner=10pt,
]{gradientbox}

\newcommand{\gradienttitle}[1]{%
  \begin{mdframed}[
    linewidth=0pt,
    innertopmargin=10pt,
    innerbottommargin=10pt,
    backgroundcolor=froggergreen,  
    fontcolor=white,
    roundcorner=0pt,
  ]
  \Large\bfseries #1
  \end{mdframed}%
}

\newcommand{\warning}[1]{{\color{red}\textbf{#1}}}

\icmltitlerunning{Frog Soup: Zero-Shot, In-Context, and Sample-Efficient Frogger Agents}

\begin{document}

\twocolumn[
\icmltitle{Frog Soup: Zero-Shot, In-Context, and Sample-Efficient Frogger Agents}

\begin{icmlauthorlist}
\icmlauthor{Xiang Li, Yiyang Hao, Doug Fulop}{inst}
\end{icmlauthorlist}
\icmlaffiliation{inst}{Stanford University}

\icmlcorrespondingauthor{Xiang Li}{kevin02@stanford.edu}
\icmlcorrespondingauthor{Yiyang Hao}{yyhao@stanford.edu}
\icmlcorrespondingauthor{Doug Fulop}{dougfulop@gmail.com}

\icmlkeywords{Reinforcement Learning, Large Language Models}

\vskip 0.3in
]

\printAffiliationsAndNotice{}

\begin{abstract}
One of the primary aspirations in reinforcement learning research is developing general-purpose agents capable of rapidly adapting to and mastering novel tasks. While RL gaming agents have mastered many Atari games, they remain slow and costly to train for each game. In this work, we demonstrate that latest reasoning LLMs with out-of-domain RL post-training can play a challenging Atari game called Frogger under a zero-shot setting. We then investigate the effect of in-context learning and the amount of reasoning effort on LLM performance. Lastly, we demonstrate a way to bootstrap traditional RL method with LLM demonstrations, which significantly improves their performance and sample efficiency. Our implementation is open sourced at \url{https://github.com/AlienKevin/frogger}.
\end{abstract}

\section{Introduction}
Atari games have proven to be valuable testing environments for RL algorithms, leading to advancements such as Deep Q Networks \cite{mnih2013playing}. RL game playing agents are traditionally initialized with zero pre-existing knowledge about a specific game environment and learn to play the game through millions of interactions with the environment. Significant time and compute is often spent exploring states that will not be experienced during high scoring policies. Exploration is particularly challenging in environments that require long horizon action sequences and provide sparse rewards, such as the Atari games and real-world robotics challenges where the state space is too large to effectively sample through free-form exploration.

In this paper we will explore whether pretrained general RL agents like reasoning LLMs can play Atari games and investigate ways to leverage pretrained RL agents to reduce the training samples for training smaller agents from scratch. We first explore whether the contextual understanding and reasoning capabilities of multi-modal LLMs extend to successful zero-shot game play. Next, we analyze the performance of latest reasoning LLMs that are underwent RL post-training in domains such as math and code. We explore a variety of methods for leveraging LLMs with zero-shot and in-context RL and investigate whether LLMs can accelerate trainings of RL agents from scratch. We will evaluate whether our approach can reduce the training time required to reach equivalent performance thresholds to traditional DQN self-play. If this approach to prioritizing exploration is successful then it may accelerate the training of RL agents for other discrete game environments, and we can extend this research to explore opportunities in continuous state space game environments (e.g. 3D action games) as well as simulators for real-world robotics applications.

\section{Background}

\subsection{RL for Atari Games}

Existing RL methods that train from scratch often require massive number of training samples, which makes them resource intensive and limits their application to real-world domains where interaction with the environment is costly. For example, the original Nature DQN \cite{mnih2015human} trained for 50 million frames to reach human-level performance.

\begin{figure*}[h]
    \centering
    \includegraphics[width=0.8\linewidth]{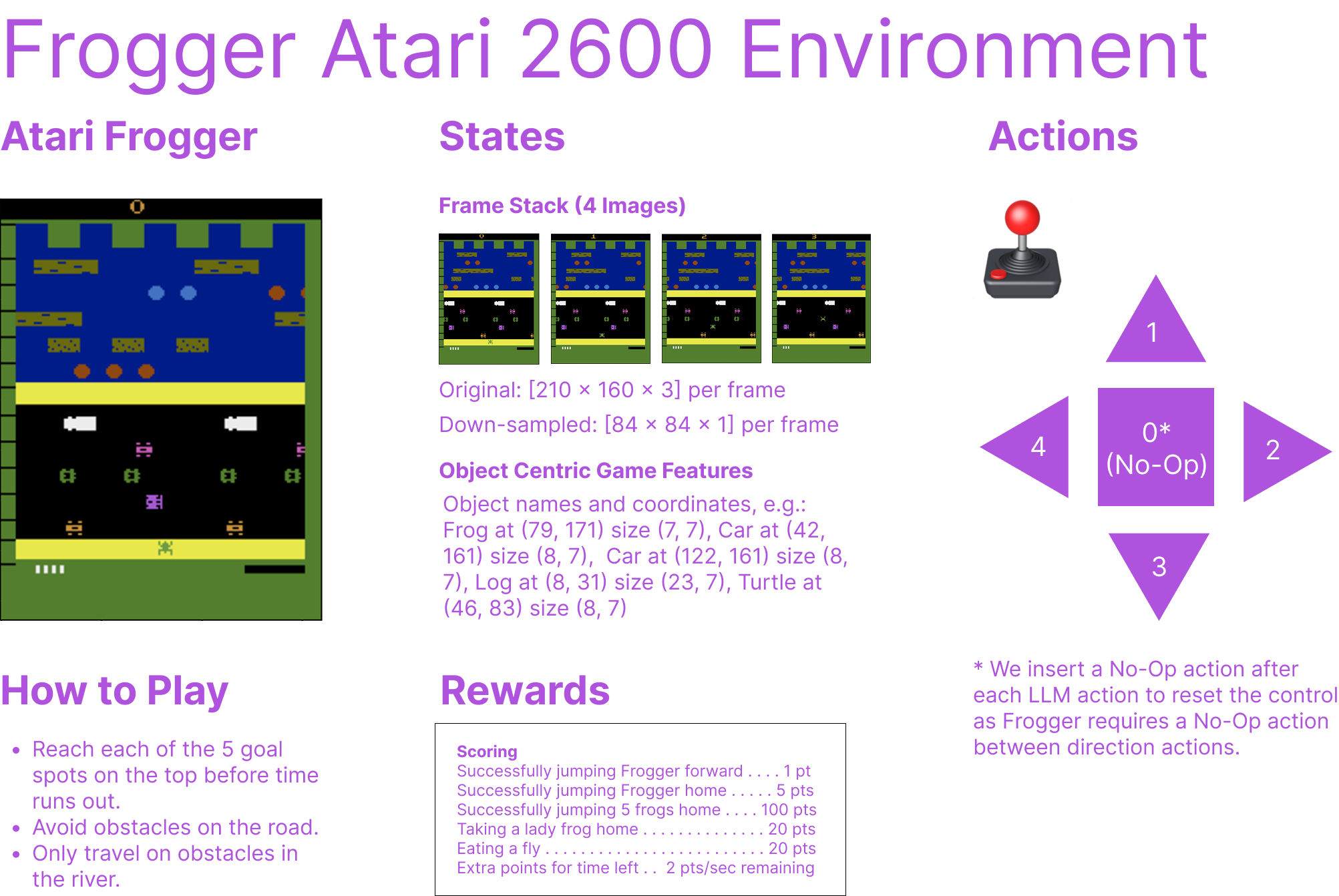}
    \caption{States, actions, and rewards of the Atari 2600 Frogger game environment}
    \label{fig:game}
\end{figure*}

\subsection{Atari Game Selection}
Traditional DQN has demonstrated strong performance in Atari games without any prior knowledge, even surpassing human players in some cases. However, certain games pose significant challenges for DQN when they include:\\
1. Sparse Rewards: Many games only provide rewards at the end of the episode, making it difficult for the agent to learn which actions contribute to success. Key actions often lack immediate feedback but are crucial for long-term success.\\
2. Exploration Challenges: Large state-action spaces make it inefficient for the agent to learn by random exploration, requiring significant time to discover meaningful trajectories.\\
3. Long Term Planning: Some games demand strategic foresight and an understanding of game mechanics across multiple phases, which is difficult for a DQN agent to learn without explicit memory or hierarchical decision-making.

While games like Montezuma's Revenge and Pitfall are famous for their sparse rewards and exploration challenges, the pace of those games are significantly slower than games like Pong and Breakout. After some search, we identified Frogger as a great mix of planning and fast-paced game play. Frogger features a multi-phase gameplay where road-crossing and river-crossing follow different rules, as players must avoid objects while crossing the road but must land on objects while crossing the river. Under a strict 30 seconds time limit and faced with fast moving cars and logs at every turn, the player has to make rapid low-level decisions while planning their future routes maticulously to not be hit by a car or fall into water. While moving the frog up gives a 1 point reward, the most significant rewards are sparse and difficult to reach since the player has to move the frog all the way through 12 lanes to reach the home bays. Additionally, the river section requires sophisticated multi-step planning to land the frog in one of the unoccupied home bays.

\subsection{Atari 2600 Frogger Game Mechanics}

The objective of the Frogger game is to hop as many frogs to safety as the player can and to score the most points along the way. The frog needs to cross a road with five lanes to reach the riverbank. Then, it needs to cross a river with five lanes dotted with logs and turtles to reach one of the home bays. When all five home bays are filled with frogs, the player receives a hefty reward of 100 points and 2 points for the number of seconds remaining to encourage fast traversals. The player is given at most 30 seconds to move a frog into the home bay and the frog automatically dies on time out. Then, the player moves on to the next level with more and faster cars. Later levels also have alligators in the river and snakes on the riverbank and logs that can bite the player. There is no limit to the number of levels and the player can theoretically earn infinite points if they don't make a mistake.

\begin{figure*}[h]
    \includegraphics[width=\linewidth]{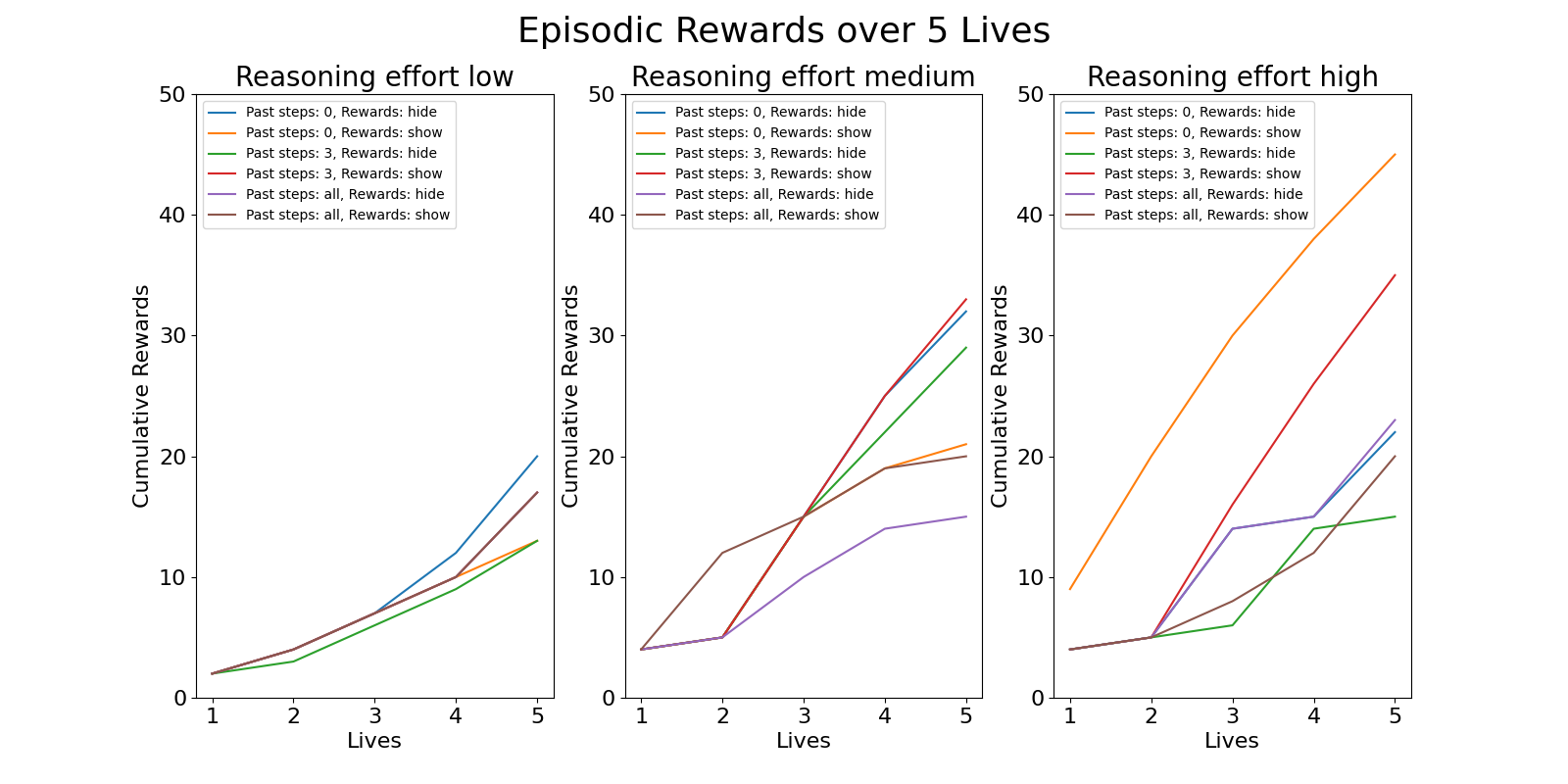}
    \caption{Zero-Shot/In-Context RL with o3-mini on Object-Centric Representation of Frogger}
    \label{fig:o3-mini}
\end{figure*}

\section{LLM Agents}
Two research questions we have are: "Can LLMs play Atari games out of the box?" and "Can LLMs learn the rule of the game through a few trial and errors". Recent research suggests that this may be possible. Atari-GPT\cite{waytowich2024atarigptinvestigatingcapabilitiesmultimodal} demonstrates that modern LLMs like GPT-4o play Atari games like Frogger. The paper highlights the LLMs' ability to understand contextual game information and provide strategic guidance, something traditional RL agents lack. However, it was shown that the LLMs struggle with low-level controls and often performs worse than traditional RL algorithms like DQN. Digging into the cause of low performance, AtariGPT broke down game playing skills into visual, spatial, strategy, and identification subtasks. GPT-4o was shown to excel in all subtasks except spatial reasoning. In particular, Multimodal LLMs were found to have trouble accurately assessing the distance between game objects.

\subsection{Spatial Reasoning of Multimodal LLM}
To further investigate whether latest Multimodal LLMs have sufficient spatial reasoning skills for playing Frogger, we tested two latest vision LLMs: gpt-4o-2024-11-20 and Claude-3.7-sonnet. To make sure that the models can see the frames clearly, we tested two frame size configurations: the original 210 by 160 pixels and an upsampled 1024 by 1024 pixels version. The upsampled version not only shows the objects more clearly, but also gives the model more visual tokens and effectively increases the amount of available inference compute. We found that models can identify gaps between the car and the frog in the next lane but have difficulty assessing the size of the gap and ended up bumping into cars by choosing to move up. See Supplement section \ref{spatial_reasoning} for a detailed case study.

\subsection{Object-Centric Atari}
To unleash the true performance of LLMs unhindered by the vision encoders, we extract key objects from the frames using the Object-Centric Atari (OCAtari) framework \cite{delfosse2024ocatariobjectcentricatari2600}. OCAtari extracts a list of key objects along with their categories, x-y locations, and their widths and heights. Notably, even though the category, locations and sizes of the objects are much easier to access under the OCAtari framework, other useful information such as the general divide between the river and the road, the location of the five home bases at the top of the frame, lives left, the time bar, and the total scores are not available to the models. Additionally, OCAtari has shown that object-centric PPO agents perform on par with the pixel-based PPO agents on Freeway, a simplified version of Frogger \cite{delfosse2024ocatariobjectcentricatari2600}. Hence, we think OCAtari presents a fair environment to compare against pixel-based agents. See the middle of Figure \ref{fig:game} for an example object-centric representation of a single frame.

\subsection{Zero-Shot LLM Agents}
Since traditional LLMs like GPT-4o have difficulty reasoning through the coordinates, we only report results with the latest reasoning models. These reasoning models are pretrained on massive amount of world knowledge during their pretraining phase and then underwent additional RL post-training to solve challenging math and coding questions. We first test o3-mini on a zero-shot setting without showing it past rewards (rewards: hide). We test three configurations under this setting, one with only the current step's state and action (past steps: 0), one that additionally includes three past steps' states and actions (past steps: 3), and one that includes all past steps (past steps: all). The 3 past steps exactly mirrors the common 4 framestack configurations used in DQN works\cite{mnih2015human} except past actions are also included in our case. For each of the configurations, we test under low, medium, or high reasoning efforts. \textbf{For the zero-shot setting, medium reasoning with 0 past steps achieves the highest episodic reward of 32, reaching as far as the 10th lanes out of the total 12 lanes in the game.}

As shown in Figure \ref{fig:o3-mini}, \textbf{including more past steps counterintuitively leads to decreased performance across levels of reasoning efforts}, evident in the fact that 0 past steps performed the best among configurations with hidden rewards (blue lines are higher in the y axis compared to green and purple lines). Based on analysis of the LLM reasoning traces, we found one potential reason for this counterintuitive result is that the \textbf{richer context adds noise and biases the LLM during its reasoning}. We found that even model with high reasoning effort are susceptible to action frequency bias and would prefer taking more risky actions like going up because it was frequently taken in the past. See Supplement section \ref{sample_llm_step} for an example prompt and response.

\begin{figure}[h!]
    \includegraphics[width=\columnwidth]{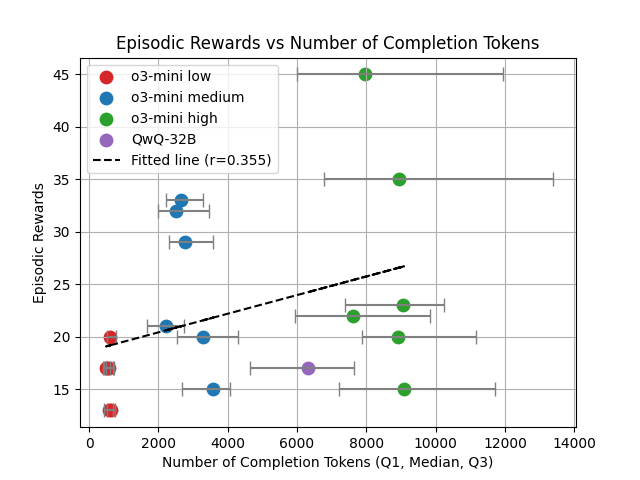}
    \caption{Episodic rewards vs number of completion tokens with o3-mini and QwQ-32B}
    \label{fig:rewards_vs_tokens}
\end{figure}

\subsubsection{Reasoning Effort and Performance}

While we noticed consistent improvement to configurations like 0 past steps with rewards shown as reasoning effort increases in Figure \ref{fig:o3-mini}, other configurations go through ups and downs. In Figure \ref{fig:rewards_vs_tokens}, we found \textbf{a weak positive correlation (r=0.355) between episodic reward and number of completion tokens}. The main reason for the weak correlation is due to the model's sensitivity to different prompting strategies where the number of past steps and presence of rewards significantly impact performance. We also notice that the variance between performance of different configurations increases as reasoning effort increases, with \textbf{high reasoning effort being the most sensitive to prompting strategies}.


\subsubsection{Zero-Shot Performance of Open-source LLM Agent}
Recently, open-source reasoning models like QwQ-32B has demonstrated comparable performance to o1 on math, coding, and instruction-following benchmarks. We experiment with the latest QwQ-32B model to investigate the performance of open-source reasoning models on playing the challenging Frogger game. Under setting with 3 past steps and hidden rewards, which is the most common setup for Atari agents, we found that \textbf{QwQ-32B scored an episodic reward of 17 points, reaching as far as the 8th lane out of the total 12 lanes}. Given the same setup, o3-mini on low reasoning effort got 13 points, medium got 29, and high got 15. With a median completion token of 6315, \textbf{QwQ-32B outputs tokens between the range of o3-mini medium and high reasoning efforts. Its performance is also between these two o3-mini reasoning efforts.} Open-source models like QwQ-32B demonstrate strong game playing capabilities on Frogger and present exciting opportunities for RL fine-tuning, which we leave to future work. See Supplement section \ref{pdf:qwen-qwq-32b_default_past_3_rewards_hide} for LLM outputs at the end of each life.

\subsection{In-Context LLM Agents}
Recent reasoning LLMs are known for their exceptional in-context learning and reasoning capabilities. To allow the LLM to better reflect on its past policies and improve its game playing strategy, we also test the three configurations above but showing rewards for each past action. We found that the usefulness of showing rewards depend heavily on level of reasoning efforts. On low reasoning effort, showing reward degrades the performance by biasing the model towards taking suboptimal and risky previous actions that earned a positive reward (like in the case of 0 past steps with low reasoning effort, the blue line is above the orange line, see Figure \ref{fig:o3-mini}). This bias is alleviated when the past context size increases from 0 to 3. On the other hand \textbf{with high reasoning effort, we notice showing past almost always help the model achieve better performance. In the case with 0 past steps, showing the rewards to the LLM in context doubled the episodic reward from 22 to 45} (See Supplement section \ref{pdf:o3-mini_high_past_0_rewards_show} for LLM outputs at the end of each life). In the case with 3 past steps, showing the rewards in context more than doubled the episodic reward from 15 to 35. The only exception is when all past steps are included, where we see a slight decrease in episodic reward. We hypothesize this is due to including rewards for all past steps increased the amount of information the model has to process and are more likely to be biased towards frequent, positive reward actions. Hence, we have shown that reasoning LLMs can learn from past rewards and improve their policy.

\subsubsection{Explorative LLM Agent}
So far, we have prompted the LLM to choose the best action at every single step. To explore the tradeoff between exploration and exploitation, we alternatively prompt the LLM to explore the environment instead (see Supplement section \ref{sample_llm_step_explore} for an example prompt and output). We modify the configuration with high reasoning, all past steps and rewards shown to use a prompt that encourages exploration. This configuration is chosen because exploration only makes sense when the agent has a memory of past states and rewards. We found that \textbf{when asked to explore instead of picking the best action, the LLM express an intention to move up towards the river area where it is less explored. However, the overall episodic reward is only 1 point higher (21 vs 20) compared to the configuration with the best action prompt.} We think the main reason for the limited performance of both configurations is current reasoning model's limitation in analyzing long action sequences. Due to this limitation, risky and suboptimal actions are repeatedly taken despite past exploration showing them as undesirable. See Supplement section \ref{pdf:o3-mini_high_past_all_rewards_show_explore} for LLM output at the end of each life. Full LLM trajectories and game play videos for all configurations are available on HuggingFace at \url{https://huggingface.co/datasets/AlienKevin/atari-frogger-traces}.

\subsubsection{Reflective LLM Agent}
Inspired by the Reflexion framework \cite{shinn2023reflexionlanguageagentsverbal} which allows LLM agents to verbally reflect on task feedback signals, store reflection in memory and improve decision-making in subsequent trials, we design a simplified Reflexion LLM agent for playing Frogger. We first let the model play a full episode of Frogger to completion. Then, the model is asked to reflect on the states, actions, and rewards collected from the past episode and provide a list of observations and advice to improve game play. Equipped with the feedback, the agent is tasked to play the game again. For the experiment, we first reflect on the trajectory sampled by the best configuration (high reasoning, 0 past steps, and rewards shown) and then append the feedback into the prompt to play the game again under the same configuration (see Supplement section \ref{reflection_prompt_and_output} for the reflection prompt and LLM output). While the LLM feedback includes accurate observations and advice about the game, \textbf{the feedback is usually high-level and lacks key information such as the objects on the current lane become invisible and important details such as the exact speed of the cars.} Following the useful advice to align the frog relative to objects on the next lanes but not being told that the objects on the current lane become invisible, the agent repeatedly bumped into cars on the same lane and dropped into the river. Hence, \textbf{the performance drops from 45 to 21 despite the additional experience included in the prompt}. See Supplement section \ref{pdf:o3-mini_high_past_all_rewards_show_round_2} for LLM outputs at the end of each life)

\subsection{Depth-First Search Agent}
Atari games run deterministically so an agent can technically memorize the game layout without learning a robust policy. To prevent this exploit, later versions of the Arcade Learning Environment implements sticky actions to inject stochasticity into the game \cite{machado2017revisiting}. Sticky actions allow the the environment to execute the agent’s previous action again, instead of the agent’s new action with some probability. To understand the complexity of the action sequence required to play a deterministic version of Frogger, we wrote a Depth First Search (DFS) algorithm that cloned the game environment at every frame to explore each of the 5 available actions. We later discovered that this technique has been applied to other Atari 2600 games under the name "The Brute" \cite{machado2017revisiting}.

For our DFS algorithm, depending on the initial frame used to initiate the search (between 110 and 140), the algorithm could achieve a score of 8 (corresponding to successfully crossing the road then traveling 2 rows across the river) within a total of 35 simulated steps and 90 seconds of real world clock time on a mid-range desktop PC. Randomizing the action selection order did not meaningful impact required exploration steps. However, achieving a score of 9 (progressing one additional block in the water) required a minimum of 1,320 additional simulation steps and 23 minutes. The search to achieve a score of 10 exceeded memory limitations on our PC in our implementation without attempt at optimization. In a default Frogger environment where the Frog has 4 lives, this suggests \textbf{a naive policy could be trivially learned that achieves a baseline score between 32 and 36. However, as shown in \cite{machado2017revisiting}, "the Brute" is very sensitive to small perturbations in the environment such as sticky actions and performance drastically decreases under more realistic game environments} with even small amount of randomness. Additionally, the DFS procedure necessitates the storage and reloading of intermediate states, which is not generally available for many realistic environments.

 \begin{figure*}[htbp]
    \centering
    \includegraphics[width=0.8\linewidth]{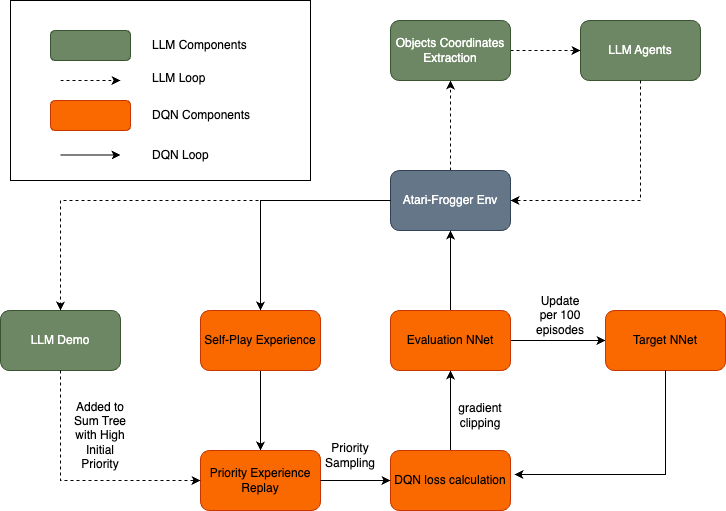}
    \caption{Enhance DQN with LLM-generated demonstrations}
    \label{fig:DQN_System}
\end{figure*}

\subsection{DQN Agent}
Our traditional RL algorithm baseline is Deep Q-Network (DQN). In our implementation, the DQN agent processes the game's visual observation using a three-layer convolutional neural network (CNN) and makes action decisions using a two-layer fully connected network.

To ensure efficient processing of game images, we apply the following standard preprocessing steps first used by \cite{mnih2015human}. First, we convert RGB frames (210x160x3) to grayscale resized images (84x84x1) to reduce conputational cost while preserving essential information. Second, we stack four consecutive frames as input, enabling agent to infer motion dynamics.

Training DQN agents on Atari Games can be computationally expensive, especially for challenging games like Frogger. To establish a benchmark, we use simplified training setup. First, we limited Training Episodes: 5000 episodes to compare traditional DQN and LLM-guided DQN within a practical timeframe. Second, the replay buffer space is constrained to 8000 total time steps (states, action, rewards, next states pair) due to RAM limitation (50GB) for Google Colab.

Inspired by DQfD \cite{hester2017deepqlearningdemonstrations}, \textbf{we incorporate LLM-generated demonstrations using Prioritized Experience Replay (PER) \cite{schaul2016prioritizedexperiencereplay}. PER modifies the experience sampling probability by assigning higher priority to important transitions.} For the standard DQN, we setup an empty Priority Replay Buffer and calculate priority value for self-generated trajectories. 

$$p_i = (|\delta_i| + \epsilon)^\alpha$$

where $\delta_i$ is the TD-error, $\epsilon$ is a small positive value ensuring that no priority is zero, and $\alpha$ decides the amount of prioritization. See Figure \ref{fig:DQN_System} for a diagram of our DQN system enhanced with LLM demonstrations.

\subsection{LLM Demonstrations Guided DQN}

In the LLM-guided DQN, the first step is to collect LLM-generated gameplay demonstrations. Based on our previous experiments with LLM agents, we chose o3-mini as our model, and implement object coordinates as inputs to guide decision-making. 

The image-to-text conversion process consists of several key steps:
\begin{itemize}
    \item Preprocessing the Game Observations. 1. Convert RGB game frames into grayscale 2D arrays of size 210×160 with uint8 pixel values. 2. Crop the image to focus only on the gameplay screen, removing irrelevant UI elements.
    \item Binary Image Segmentation. 1. Convert the grayscale image into a binary representation (foreground objects vs. background). 2. Identify objects and apply filters to refine segmentation. Since some vehicles consist of multiple disconnected components (e.g., wheels separated from the main body), we filter out objects smaller than 5 pixels to remove noise.
    \item Object Detection \& Feature Extraction. For each detected object, compute its center position, width, and height. Apply a specialized frog detection algorithm, since the frog is unique in the game. It is identified as a 7×7 square with a grayscale value of 124.
    \item Final Output: Object List \& Frog Position. After processing, we obtain a structured list of objects with their bounding box coordinates and classification, along with the frog’s precise location.
\end{itemize}


we collect expert demonstrations from LLM-generated trajectories. Specifically, we use 10 LLM-played episodes, comprising a total of 1657 time steps $(s, a, r, s', d)$, which accounts for $20.7\%$ of the prioritized replay buffer. The episodic rewards range from 12 to 35.

With a prioritized replay buffer of size 8000, we initialize the buffer by preloading all expert transitions $(s, a, r, s', d)$ with initialized high priority value of 5.0, ensuring it is sampled frequently during the early training. Over the course of training, the priorities of these expert samples are dynamically updated based on TD-error as self-experience trajectories and gradually decreases as the network learns from these samples. Once the expert transitions' priorities become sufficiently low, they are naturally replaced by newly generated agent experiences.

\begin{figure}[h!]
    \includegraphics[width=\linewidth]{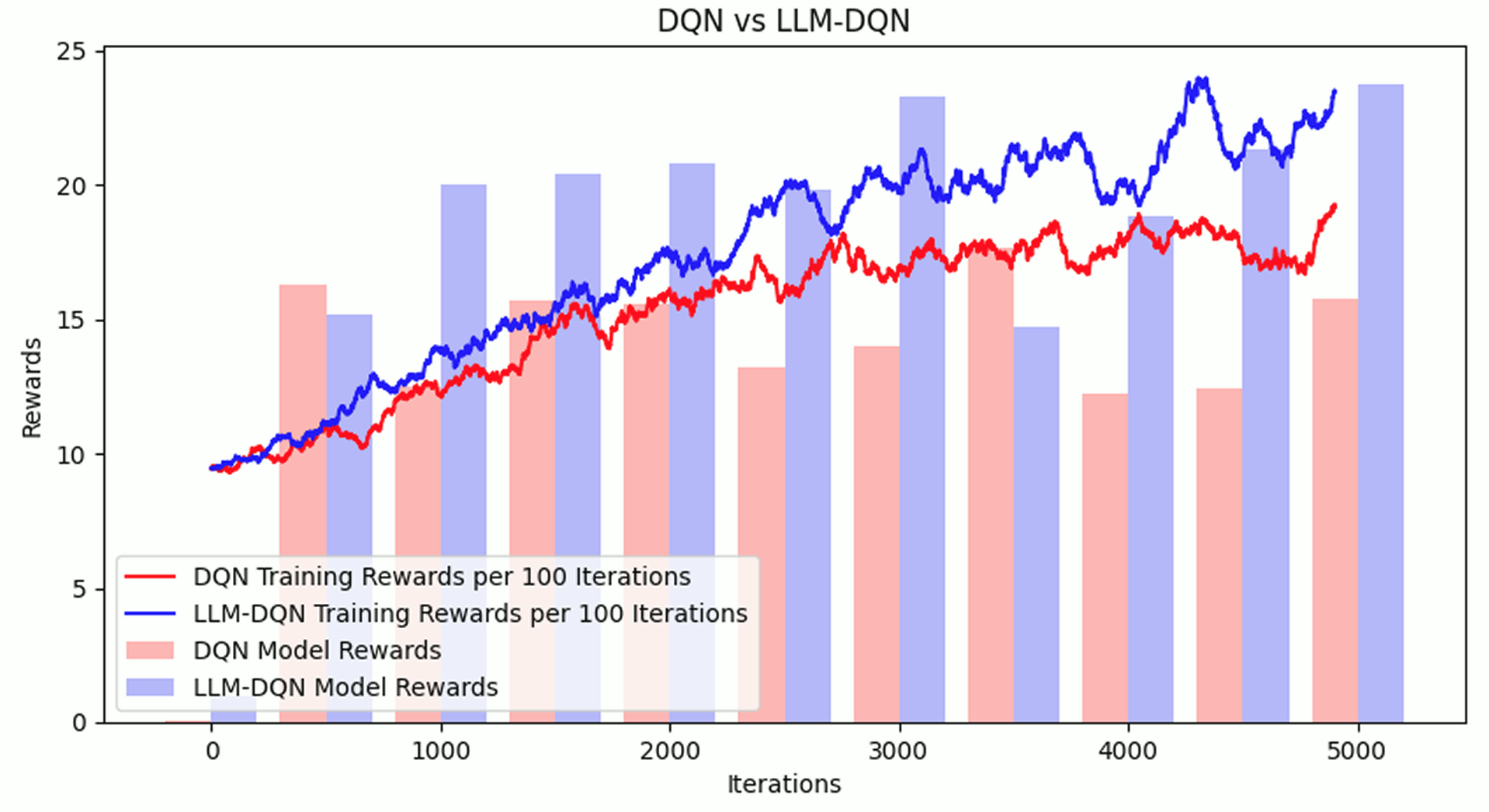}
    \caption{LLM-Guided DQN}
    \label{fig:DQNvsLLM}
\end{figure}

\subsection{Experiment results of LLM guided DQN}
Over the course of 5,000 episodes, we observed a clear upward trend in both agents' average rewards, indicating progressive learning. 

Initially, when playing randomly, the average reward was 8-10, meaning that the frog typically took only one or two steps forward before failing. Given the scoring mechanism (+1 point per step forward with 5 lives per game), this suggests that the frog struggled to make meaningful moves.

By the end of training, both agents demonstrated improved performance.
standard DQN shows average 15 points which means the frog consistently advanced average two steps or more, mostly remaining on the roadside, rarely frog is able to across road and reach the river. And LLM-guided DQN shows average rewards of 24, which means frog can manage to cross road 2-3 times per episode and died falling into the river. See Figure \ref{fig:DQNvsLLM} for an example run of DQN vs LLM-Guided DQN.

\begin{figure}[h!]
    \centering
    \includegraphics[width=0.5\linewidth]{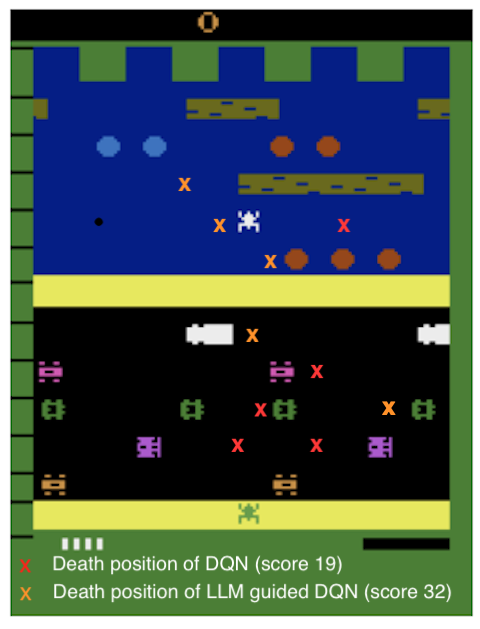}
    \caption{Example of one episode by DQN vs LLM-Guided DQN}
    \label{fig:DQNvsLLM}
\end{figure}

\textbf{The LLM-guided DQN consistently outperformed the standard DQN in total rewards throughout training.} The performance gap emerged before 1,000 episodes and continued to widen as training progressed. While traditional DQN may eventually catch up given sufficient training time, our results demonstrate that LLM-based demonstrations accelerate learning, enabling the agent to achieve higher rewards within the same number of training episodes. \textbf{LLM-Guided DQN achieved 35.3\% higher rewards after 5,000 episodes compared to standard DQN and demonstrated faster convergence.} This highlights the potential of leveraging LLM-generated demonstrations to improve the sample efficiency of RL agents. Especially at the beginning of training, when traditional RL agents struggled with random actions, LLM demonstration can show a good starting point.

\subsection{Future Research Directions}

\subsection{RL fine-tuning of Reasoning LLMs}

Since small open-source reasoning models like QwQ-32B have demonstrated strong potential on zero-shot game play, we want to explore their capabilities under more configurations, including rewards in context and as part of a reflective agent. We are most excited about the potential to fine-tune such reasoning models on the game and subsequently distill the fine-tuned model into much smaller student models for efficient inference.

\subsubsection{LLM-Guided Monte Carlo Tree Search}
\begin{figure}[h!]
    \includegraphics[width=\linewidth]{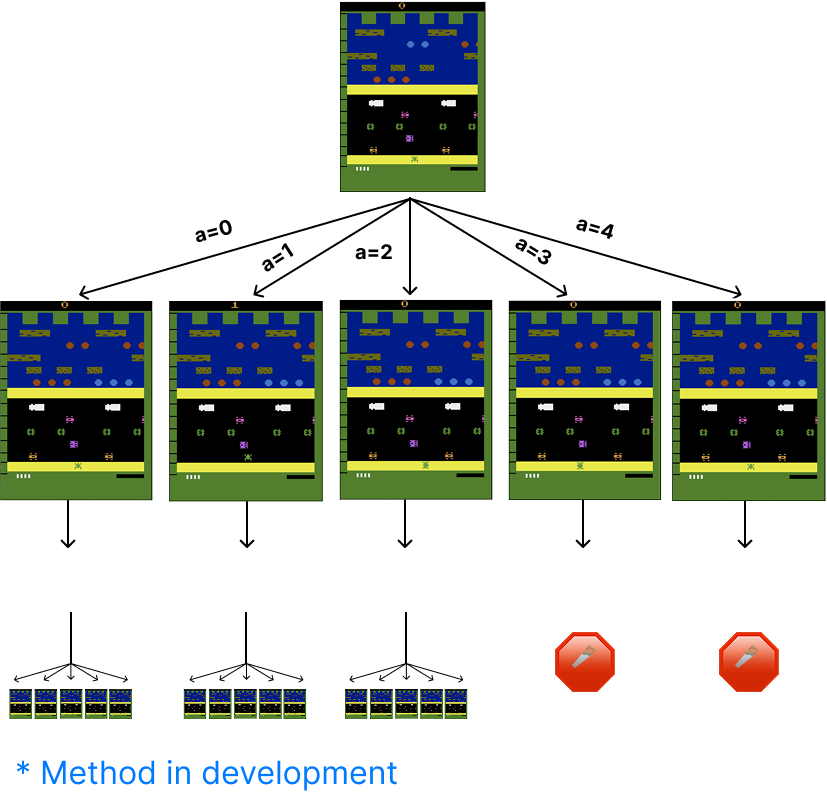}
    \caption{Monte-Carlo Tree Search with LLM Branch Pruning}
    \label{fig:mcts}
\end{figure}

As we demonstrated with our depth-first search baseline, over half of a level in Frogger can be completed through a recursive search of the state space. However, higher scoring game states remain challenging to achieve, especially when the environment is stochastic. As further research we would implement MCTS and use LLMs to strategically explore the space and prune branches in the tree, which we believe may enable an RL-trained agent to complete the first level for the first time. See Figure \ref{fig:mcts} for an illustration.

\section{Conclusion}

In this work, we have demonstrated that reasoning LLMs can achieve reasonable performance on the challenging Frogger game when given only the previous states, actions, and rewards. We found that increased reasoning effort correlates with LLM performance but models are sensitive to the number of past steps included in the prompt. On high reasoning effort, having access to past rewards more than doubled model performance. Expanding beyond best next action prediction, we experiment with explorative and reflective LLM agents and found that further research is needed to alleviate bias and balance exploitation and exploration. Lastly, we show traditional RL algorithms like DQN can significantly benefit from LLM demonstrations, achieving 35.3\% higher episodic rewards under similar compute budgets.

\bibliography{paper}
\bibliographystyle{icml2025}

\onecolumn
\section{Supplements}
\subsection{Spatial Reasoning Case Studies with GPT-4o and Claude 3.7 Sonnet on Frogger} \label{spatial_reasoning}

\begin{gradientbox}
\gradienttitle{gpt-4o-2024-11-20 (210x160)}

\begin{figure}[H]
\centering
\begin{subfigure}{0.45\textwidth}
\centering
\includegraphics[width=\textwidth]{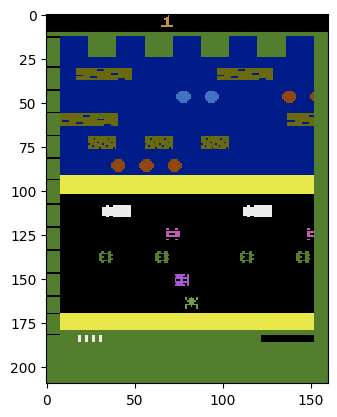}
\caption{Step: 109}
\end{subfigure}
\hfill
\begin{subfigure}{0.45\textwidth}
\centering
\includegraphics[width=\textwidth]{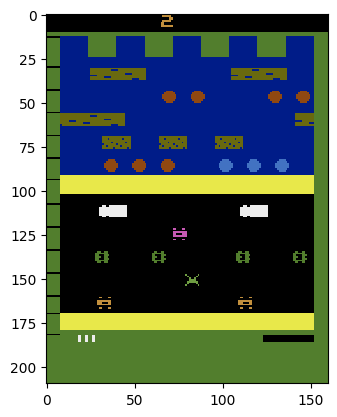}
\caption{Step 110}
\end{subfigure}
\caption{gpt-4o-2024-11-20 (210x160) gameplay frames}
\end{figure}

\noindent\textbf{Step 110 LLM output:}\\

\{\\
"game\_state":"The player frog is slightly above the white line in the game area, situated among moving logs and platforms in a challenging play zone.",\\
"reasoning":"Considering the player's position and the direction required to proceed towards the goal, \warning{moving UP aligns best to progress while avoiding potential obstacles}.",\\
\warning{"action":"UP"}\\
\}
\end{gradientbox}

\pagebreak

\begin{gradientbox}
\gradienttitle{gpt-4o-2024-11-20 (1024x1024)}
\begin{figure}[H]
\centering
\begin{subfigure}{0.45\textwidth}
\centering
\includegraphics[width=\textwidth]{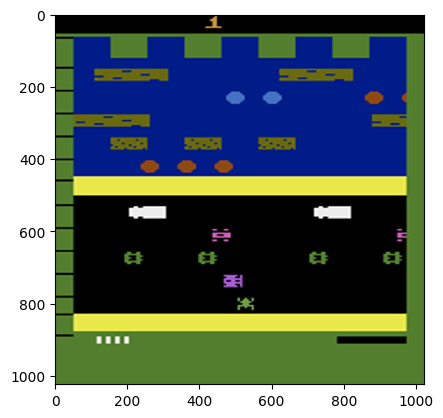}
\caption{Step: 109}
\end{subfigure}
\hfill
\begin{subfigure}{0.45\textwidth}
\centering
\includegraphics[width=\textwidth]{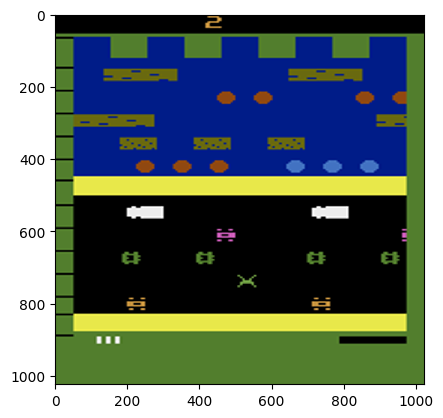}
\caption{Step: 110}
\end{subfigure}
\caption{gpt-4o-2024-11-20 (1024x1024) gameplay frames}
\end{figure}

\noindent\textbf{Step 109 LLM output:}\\
\{\\
"game\_state":"The player character is near the bottom of the screen, on a black horizontal section indicating the road area, surrounded by moving vehicles. Above this section are the river sections.",\\
"reasoning":"\warning{To advance and stay safe, an upward action might be effective if the road above is clear from obstacles and vehicles.}",\\
\warning{"action":"UP"}\\
\}
\end{gradientbox}

\pagebreak

\begin{gradientbox}
\gradienttitle{Claude-3.7-sonnet (210x160)}

\begin{figure}[H]
\centering
\begin{subfigure}{0.45\textwidth}
\centering
\includegraphics[width=\textwidth]{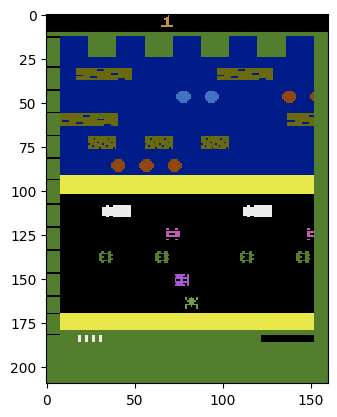}
\caption{Step: 109}
\end{subfigure}
\hfill
\begin{subfigure}{0.45\textwidth}
\centering
\includegraphics[width=\textwidth]{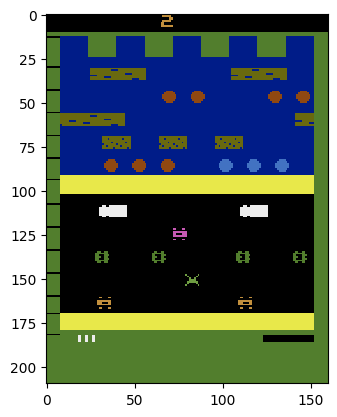}
\caption{Step: 110}
\end{subfigure}
\caption{Claude-3.7-sonnet (210x160) gameplay frames}
\end{figure}

\noindent\textbf{Step 110 LLM output:}\\
\{\\
  "game\_state": "The game appears to be Frogger or a similar game. The screen is divided into different zones. The upper area has a blue background (likely representing water) with several moving platforms or logs indicated by green/brown rectangular objects. The bottom half has a black background (likely representing a road) with what appear to be vehicles (white and green objects) moving across. The player character (small purple object) is currently positioned in the road area near the bottom of the screen. There are white markings at the very bottom which likely represent destinations or safe zones.",\\
  "reasoning": "The player character is currently in the road section with vehicles moving horizontally. Looking at the positioning, moving UP would be the logical choice as the goal in Frogger-like games is typically to cross from the bottom to the top while avoiding obstacles. Moving UP would progress the character toward the water section where the player would need to hop on the moving platforms to continue advancing. Moving LEFT or RIGHT would likely keep the character in the dangerous road section, while moving DOWN would move away from the goal. \warning{Since UP was rewarded with 1.0 in the previous step, continuing in that direction seems to be the correct strategy}.",\\
  \warning{"action": "UP"}\\
\}
\end{gradientbox}

\pagebreak

\begin{gradientbox}
\gradienttitle{Claude-3.7-sonnet 1024x1024}

\begin{figure}[H]
\centering
\begin{subfigure}{0.45\textwidth}
\centering
\includegraphics[width=\textwidth]{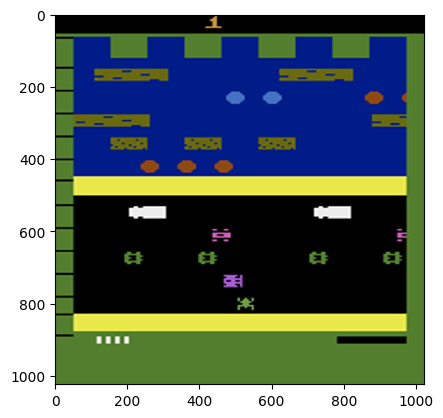}
\caption{Step: 109}
\end{subfigure}
\hfill
\begin{subfigure}{0.45\textwidth}
\centering
\includegraphics[width=\textwidth]{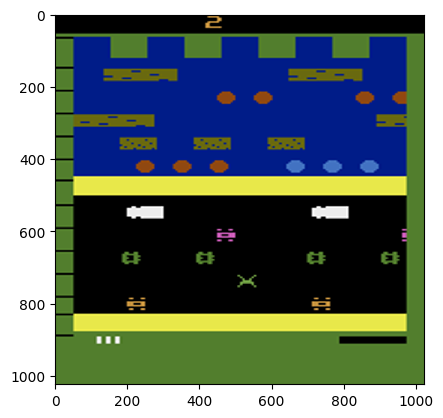}
\caption{Step: 110}
\end{subfigure}
\caption{Claude-3.7-sonnet (1024x1024) gameplay frames}
\end{figure}

\noindent\textbf{Step 109 LLM output:}\\
\{\\
  "game\_state": "This appears to be a Frogger-like game. The player controls a small green character (likely a frog) at the bottom of the screen. The game is divided into two sections: the lower black section appears to be a road with various vehicles (pink and green objects representing cars/trucks), and the upper blue section appears to be water with logs or platforms (yellow/brown objects). The player needs to navigate across these hazards. Currently, the player is positioned near the bottom of the screen in the road section with vehicles approaching from both sides. There's a numeric display showing '1' at the top, likely indicating the level or remaining lives.",\\
  "reasoning": "The player needs to safely navigate through the traffic on the road section before reaching the water section. Looking at the current position, there are several vehicles nearby, particularly pink vehicles above and green vehicles to the sides. \warning{Moving UP seems to be the safest immediate option as there appears to be a gap in the traffic directly above the player.} Moving LEFT or RIGHT would risk collision with the green vehicles. Moving DOWN would bring the player closer to the bottom edge with no apparent advantage. Since you received a reward of 1.0 for the UP action in the previous step, continuing to move upward through traffic gaps appears to be the successful strategy.",\\
  \warning{"action": "UP"}\\
\}
\end{gradientbox}
\pagebreak

\subsection{Sample step of LLM run on Frogger}\label{sample_llm_step}

\begin{gradientbox}
\gradienttitle{Example step with reasoning=low, past steps=0 and rewards=show}
\textbf{LLM Prompt:}
Step: 55, action: NOOP, reward: 0, game objects: Frog at (79, 171) size (7, 7), Car at (36, 161) size (8, 7), Car at (116, 161) size (8, 7), Car at (72, 148) size (8, 7), Car at (63, 135) size (8, 7), Car at (31, 135) size (8, 7), Car at (143, 135) size (8, 7), Car at (111, 135) size (8, 7), Car at (65, 122) size (8, 7), Car at (145, 122) size (7, 7), Car at (113, 109) size (16, 7), Car at (33, 109) size (16, 7), Log at (82, 70) size (16, 7), Log at (50, 70) size (16, 7), Log at (18, 70) size (16, 7), Log at (134, 57) size (18, 7), Log at (8, 57) size (30, 7), Log at (91, 31) size (32, 7), Log at (11, 31) size (32, 7), Turtle at (40, 83) size (8, 7), Turtle at (72, 83) size (8, 7), Turtle at (56, 83) size (8, 7), Turtle at (140, 44) size (8, 7), Turtle at (96, 44) size (8, 7), Turtle at (80, 44) size (8, 7)
\\\\
You are an expert gamer and your goal is to choose the best action to beat the game.

The game objects are given by their top-left corner's (x, y) positions followed by their width and height in (w, h).

Think about all possible actions and why each action is or is not the best action to take. You are at step 55 and the potential actions you can take are NOOP, UP, RIGHT, LEFT, DOWN.

Output in this JSON format: \{game\_state: describe the current game state in detail, reasoning: reasoning for choosing an action, action: the chosen action\}

\textbf{LLM response:}
\{\\
    "game\_state: "At step 58, the frog is at (79,148) with size (8,7), meaning it has successfully advanced upward from the starting area. The current objects on the screen include several cars, logs, and turtles in distinct lanes. The cars are present in lanes covering vertical positions around 161, 135, and 109, while logs and turtles are further up (rows 70 and 44 respectively). The frog is in a lane likely corresponding to the lower part of the road where cars are moving, but its recent upward moves have earned rewards, suggesting that moving up is beneficial. The positions of the vehicles still provide a window to progress, and there is no immediate collision threat given the frog's current position.",\\
    "reasoning": "Since the frog has been rewarded for moving upward and is safely advancing along the road layer while avoiding vehicles, the best action is to keep progressing upward. The upcoming lane with vehicles or water obstacles will need careful navigation later, but for now advancing upward increases the chance of reaching the goal. Other actions like LEFT or RIGHT might not guarantee further progress, and moving DOWN would regress. Therefore, the optimal move is to press UP again.",\\
    "action": "UP"\\
\}
\end{gradientbox}

\pagebreak

\subsection{Sample step of LLM run on Frogger when instructed to explore the environment}\label{sample_llm_step_explore}

\begin{gradientbox}
\gradienttitle{Example step with reasoning=high, past steps=all and rewards=show, explore=True}
\textbf{LLM Prompt (middle steps omitted for brevity):}
Step: 2, action: NOOP, reward: 0, game objects: Frog at (79, 171) size (7, 7), Car at (142, 161) size (8, 7), Car at (62, 161) size (8, 7), Car at (99, 148) size (8, 7), Car at (19, 148) size (8, 7), Car at (116, 135) size (8, 7), Car at (84, 135) size (8, 7), Car at (36, 135) size (8, 7), Car at (8, 135) size (4, 7), Car at (119, 122) size (8, 7), Car at (39, 122) size (8, 7), Car at (8, 109) size (14, 7), Car at (86, 109) size (16, 7), Log at (126, 70) size (16, 7), Log at (30, 70) size (16, 7), Log at (8, 70) size (6, 7), Log at (28, 57) size (64, 7), Log at (119, 31) size (32, 7), Log at (39, 31) size (32, 7), Turtle at (146, 83) size (6, 7), Turtle at (18, 83) size (8, 7), Turtle at (8, 84) size (2, 5), Turtle at (83, 83) size (8, 7), Turtle at (67, 83) size (8, 7), Turtle at (51, 83) size (8, 7), Turtle at (48, 44) size (8, 7), Turtle at (32, 44) size (8, 7)\\
...\\
Step: 56, action: UP, reward: 0.0, game objects: Frog at (79, 171) size (7, 7), Car at (34, 161) size (8, 7), Car at (114, 161) size (8, 7), Car at (73, 148) size (8, 7), Car at (62, 135) size (8, 7), Car at (30, 135) size (8, 7), Car at (142, 135) size (8, 7), Car at (110, 135) size (8, 7), Car at (67, 122) size (8, 7), Car at (147, 122) size (5, 7), Car at (112, 109) size (16, 7), Car at (32, 109) size (16, 7), Log at (86, 70) size (16, 7), Log at (54, 70) size (16, 7), Log at (22, 70) size (16, 7), Log at (136, 57) size (16, 7), Log at (8, 57) size (32, 7), Log at (95, 31) size (32, 7), Log at (15, 31) size (32, 7), Turtle at (38, 83) size (8, 7), Turtle at (70, 83) size (8, 7), Turtle at (54, 83) size (8, 7), Turtle at (136, 44) size (8, 7), Turtle at (92, 44) size (8, 7), Turtle at (76, 44) size (8, 7)\\
\\
\textbf{You are a curious gamer and your goal is to explore the game environment.}\\
The game objects are given by their top-left corner's (x, y) positions followed by their width and height in (w, h).\\
You are at step 56 and the potential actions you can take are NOOP, UP, RIGHT, LEFT, DOWN.\\
Output in this JSON format: \{game\_state: describe the current game state in detail, reasoning: reasoning for choosing an action, action: the chosen action\}

\textbf{LLM response:}\\
\{\\
    "game\_state: "At step 58, the frog is at (79,148) with size (8,7), meaning it has successfully advanced upward from the starting area. The current objects on the screen include several cars, logs, and turtles in distinct lanes. The cars are present in lanes covering vertical positions around 161, 135, and 109, while logs and turtles are further up (rows 70 and 44 respectively). The frog is in a lane likely corresponding to the lower part of the road where cars are moving, but its recent upward moves have earned rewards, suggesting that moving up is beneficial. The positions of the vehicles still provide a window to progress, and there is no immediate collision threat given the frog's current position.",\\
    "reasoning": "Since the frog has been rewarded for moving upward and is safely advancing along the road layer while avoiding vehicles, the best action is to keep progressing upward. The upcoming lane with vehicles or water obstacles will need careful navigation later, but for now advancing upward increases the chance of reaching the goal. Other actions like LEFT or RIGHT might not guarantee further progress, and moving DOWN would regress. Therefore, the optimal move is to press UP again.",\\
    "action": "UP"\\
\}
\end{gradientbox}

\subsection{Samples from o3-mini-2025-01-31 with high reasoning, 0 past steps, and rewards shown}
\label{pdf:o3-mini_high_past_0_rewards_show}
\includegraphics[width=\textwidth, trim=0 100 0 0, clip]{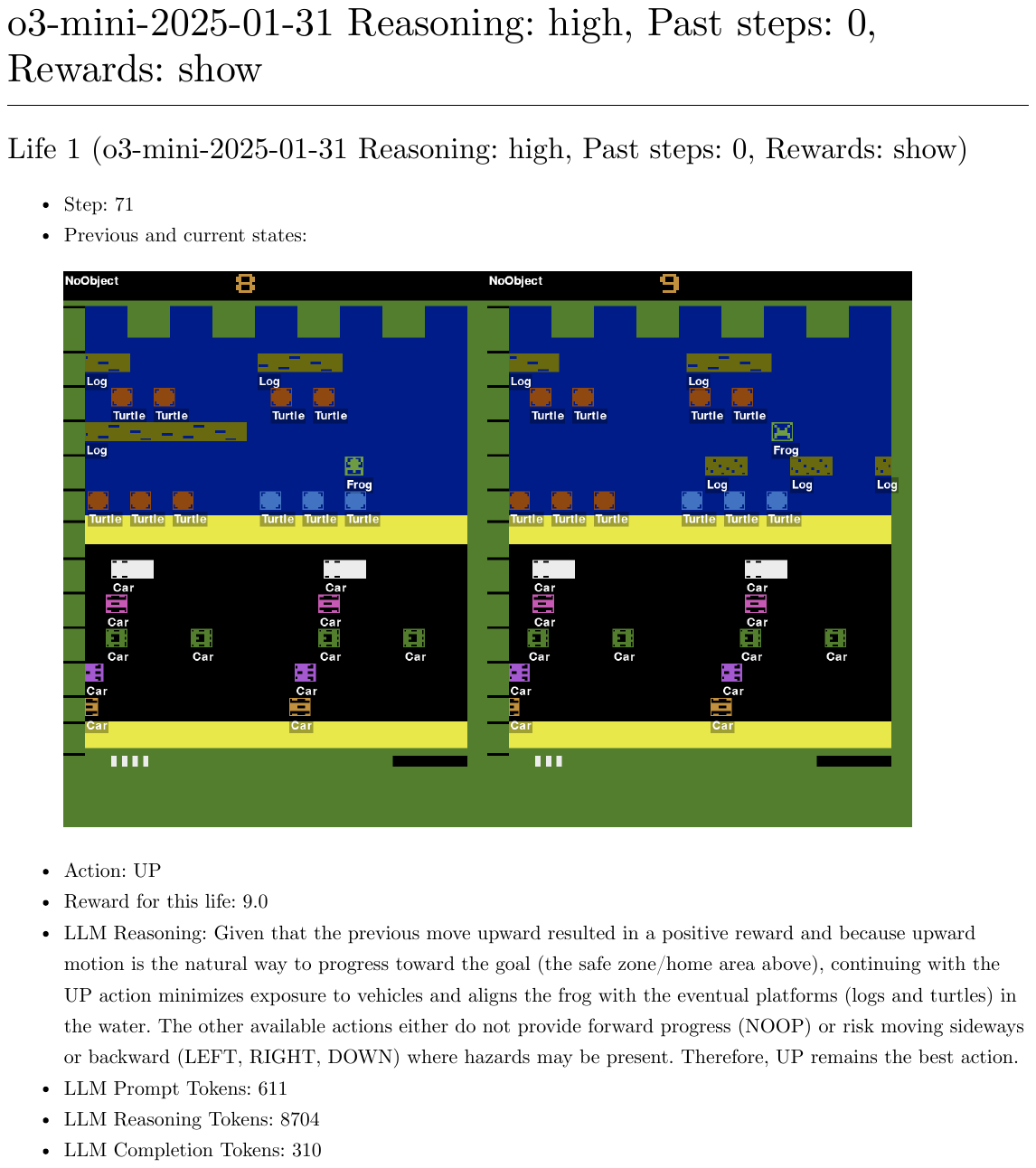}
\clearpage
\includepdf[pages=2-, scale=0.85, pagecommand={}]{o3-mini_high_past_0_rewards_show.pdf}

\subsection{Samples from o3-mini-2025-01-31 with high reasoning, all past steps, rewards shown, and explore prompt}
\label{pdf:o3-mini_high_past_all_rewards_show_explore}
\includegraphics[width=\textwidth, trim=0 100 0 0, clip]{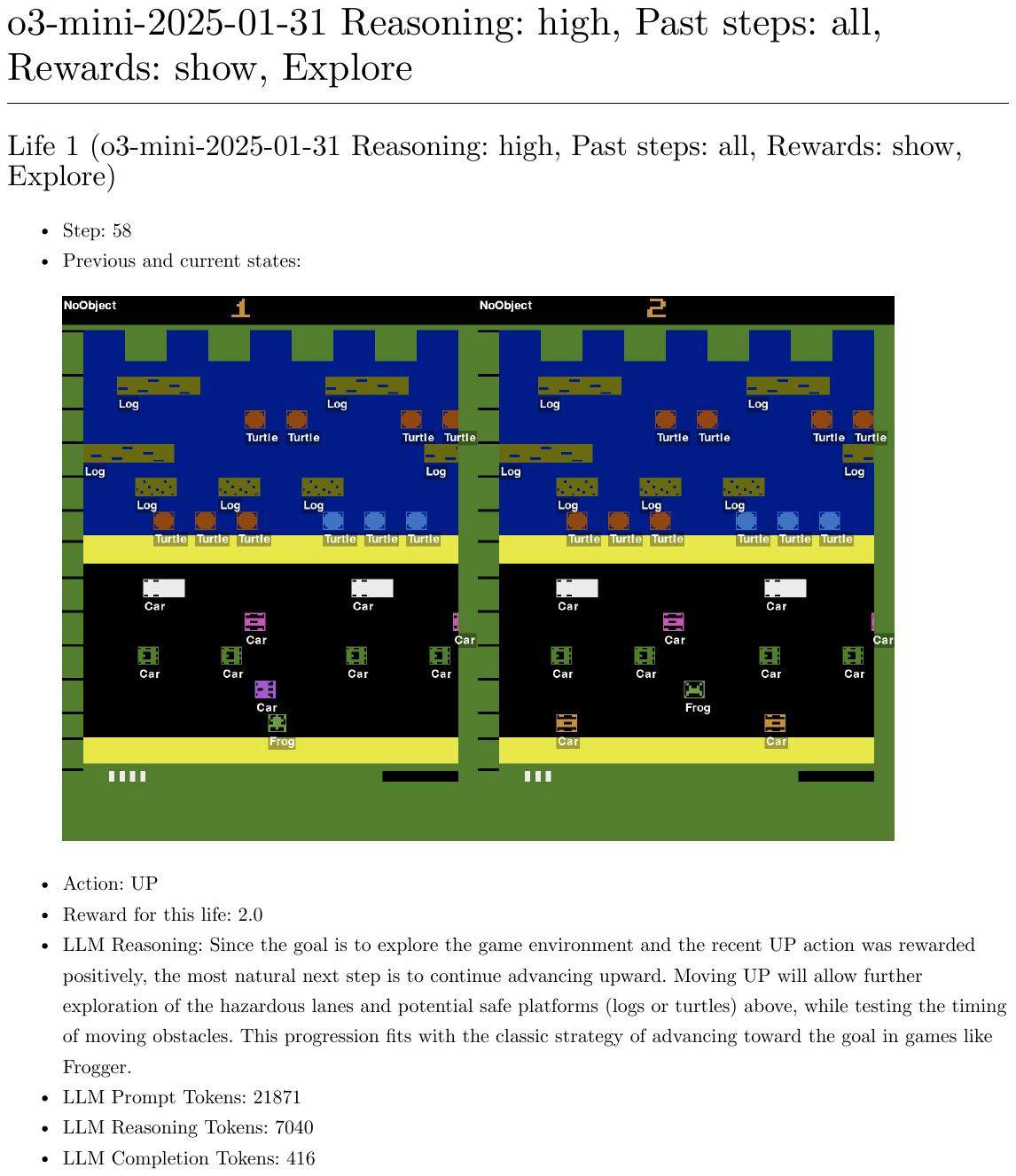}
\clearpage
\includepdf[pages=2-, scale=0.85, pagecommand={}]{o3-mini_high_past_all_rewards_show_explore.pdf}

\subsection{Samples from qwen-qwq-32b with default reasoning, 3 past steps, rewards hidden}
\label{pdf:qwen-qwq-32b_default_past_3_rewards_hide}
\includegraphics[width=\textwidth, trim=0 100 0 0, clip]{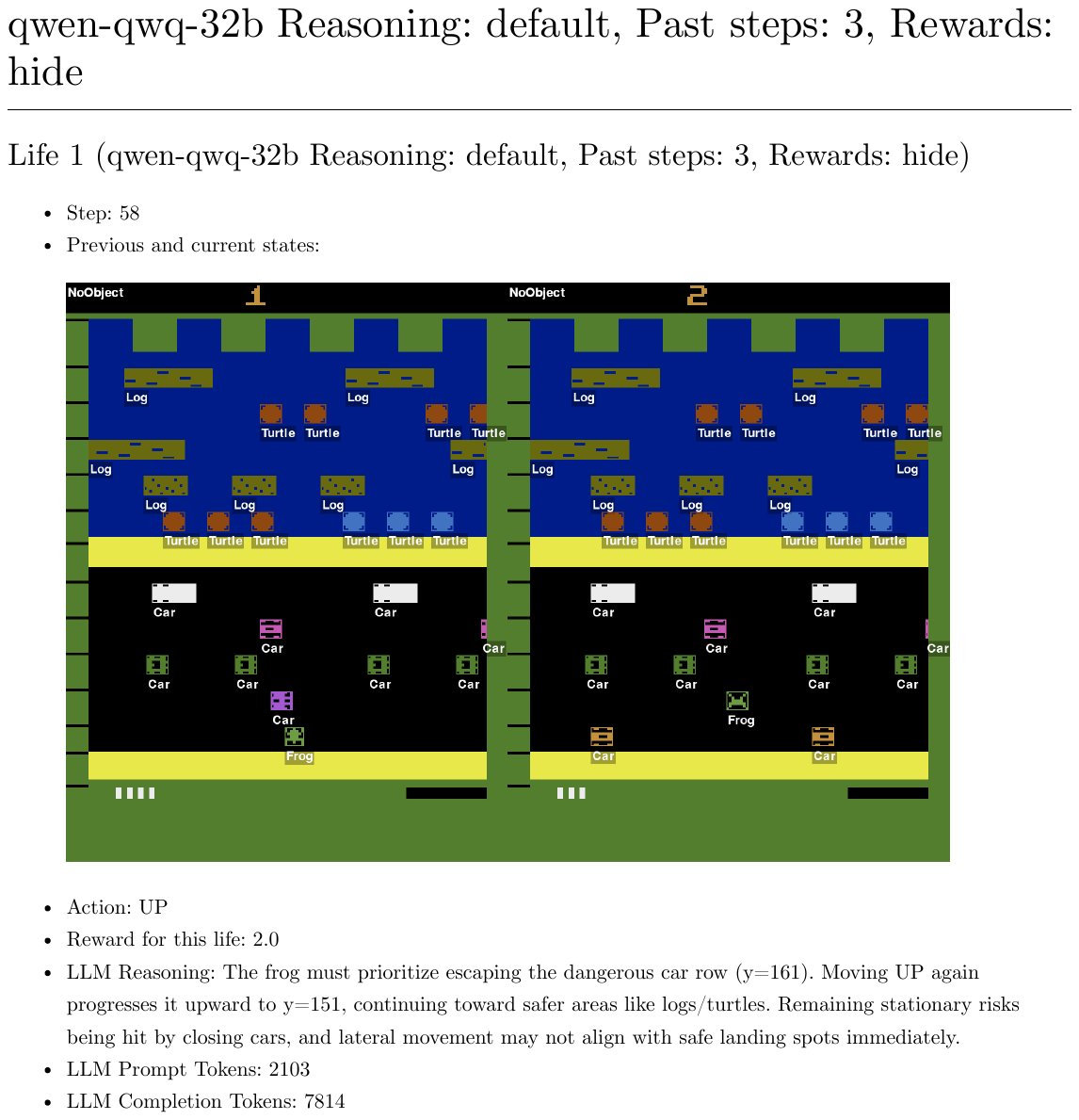}
\clearpage
\includepdf[pages=2-, scale=0.85, pagecommand={}]{qwen-qwq-32b_default_past_3_rewards_hide.pdf}

\subsection{Reflection prompt and output}\label{reflection_prompt_and_output}
\begin{gradientbox}
    \gradienttitle{Reflection prompt and output (middle steps from previous episode are omitted for brevity)}
    \textbf{Developer message:}
    The potential actions are NOOP, UP, RIGHT, LEFT, DOWN. Analyze the given episode of a game play in detail. \textbf{Provide suggestions on how to improve future game play. Be as specific in your observations and advise as possible to help the gamer maximize cumulative reward over the entire game.}

    \textbf{User message:}
    Step: 0, action: 0, reward: 0, game objects: Frog at (79, 171) size (7, 7), Car at (142, 161) size (8, 7), Car at (62, 161) size (8, 7), Car at (99, 148) size (8, 7), Car at (19, 148) size (8, 7), Car at (116, 135) size (8, 7), Car at (84, 135) size (8, 7), Car at (36, 135) size (8, 7), Car at (8, 135) size (4, 7), Car at (119, 122) size (8, 7), Car at (39, 122) size (8, 7), Car at (8, 109) size (14, 7), Car at (86, 109) size (16, 7), Log at (126, 70) size (16, 7), Log at (30, 70) size (16, 7), Log at (8, 70) size (6, 7), Log at (28, 57) size (64, 7), Log at (119, 31) size (32, 7), Log at (39, 31) size (32, 7), Turtle at (146, 83) size (6, 7), Turtle at (18, 83) size (8, 7), Turtle at (8, 84) size (2, 5), Turtle at (83, 83) size (8, 7), Turtle at (67, 83) size (8, 7), Turtle at (51, 83) size (8, 7), Turtle at (48, 44) size (8, 7), Turtle at (32, 44) size (8, 7)\\
    ...\\
    Step: 151, reasoning: \{
"game\_state": "The frog is currently at (79, 95) on the safe bank after dodging the car lanes, which lie below. The water zone above contains moving platforms – a row of turtles at y = 83 (including ones at x = 71 and x = 87) and logs further up. The frog’s x‐position at 79 leaves it caught between these platforms, risking an unsafe landing if it simply moves UP.",
"reasoning": "To progress safely into the water zone, the frog must land on one of the moving platforms. Moving UP immediately would leave the frog misaligned relative to the available turtles. By moving LEFT, the frog’s new position will be (72, 95), which aligns its horizontal range (72 to 79) neatly with the turtle at (71, 83). This repositioning sets up a safe upward move on the next step.",
"action": "LEFT"
\}, action: 3, reward: 0.0, game objects: Frog at (72, 95) size (7, 7), Car at (80, 161) size (8, 7), Car at (90, 148) size (8, 7), Car at (10, 148) size (8, 7), Car at (125, 135) size (8, 7), Car at (93, 135) size (8, 7), Car at (45, 135) size (8, 7), Car at (13, 135) size (8, 7), Car at (101, 122) size (8, 7), Car at (21, 122) size (8, 7), Car at (95, 109) size (16, 7), Car at (15, 109) size (16, 7), Log at (10, 57) size (64, 7), Log at (83, 31) size (32, 7), Log at (8, 31) size (27, 7), Turtle at (36, 83) size (8, 7), Turtle at (20, 83) size (8, 7), Turtle at (8, 83) size (4, 7), Turtle at (84, 44) size (8, 7), Turtle at (68, 44) size (8, 7), Turtle at (24, 44) size (8, 7), Turtle at (8, 44) size (8, 7), LadyFrog at (93, 70) size (7, 7)\\\\
    \textbf{LLM output:}\\
\{\\
  "reasoning": "This episode is a classic example of navigating a Frogger‐like environment where the player must carefully time upward movements to cross road lanes with moving cars and then align with safe water platforms (logs and turtles) in order to reach the final safe zone (often indicated by a LadyFrog). The agent’s decisions show that upward movement is rewarded when the frog is in a safe gap, but the frog’s horizontal positioning is crucial. Whenever there is a risk of collision with a car (or misalignment with a log/turtle) the agent appropriately chooses to adjust using LEFT or RIGHT before advancing. In several instances, the agent noted that moving directly UP would cause overlap with a car’s hitbox or would leave it misaligned from a safe platform, and so it preemptively shifted horizontally to ensure a safe jump. Overall, the gameplay was successful when rewards were given (often 1.0 for a safe upward move) and zero reward for actions that did not yield progress or resulted in risky positioning.",\\
  "observations": [\\
    "The frog starts in a safe zone and initially waits (NOOP) or moves upward repeatedly until obstacles come into play.",
    "Upward moves are the primary driver of progress and, when well timed, yield positive rewards.",\\
    "The agent often uses RIGHT moves when the frog’s left portion would potentially collide with a car or when its horizontal alignment with a log or turtle is inadequate.",\\
    "Similarly, LEFT moves are chosen when the frog’s position is too far right or if the target (LadyFrog) is on the left side, helping to realign with safe platforms.",\\
    "The agent’s reasoning identifies the layout by dividing the screen into lanes: the lower car lanes, the middle road zone, and the upper river zones with moving logs and turtles.",\\
    "There is clear evidence that when the frog is correctly positioned in safe gaps (for example, the identified vertical gaps between cars) an UP move is rewarded, while a poor alignment results in a zero reward.",\\
    "Towards the top of the episode, the frog must also align lateral position with the stationary target (LadyFrog), and the agent uses LEFT adjustments to gradually close the horizontal gap."\\
  ],\\
  "advices": [\\
    "Plan Ahead: Before moving upward, always check whether your current horizontal position ensures complete overlap with a safe gap. If part of your hitbox would overlap with a car or miss a log/turtle, delay the upward move by adjusting horizontally first.",\\
    "Use Lateral Moves Proactively: If you identify that future obstacles (cars or water hazards) can be avoided by slight shifts in the x-direction, use RIGHT or LEFT moves before entering those lanes. This preemptive repositioning is key to maintaining a safe trajectory.",\\
    "Timing is Critical: Observe the periodic movement and spacing of vehicles and platforms. Try to delay an upward move if a car is about to enter your gap, and consider using NOOP if the traffic is temporarily favorable.",\\
    "Optimize Alignment for Water Crossing: Once you finish the road portion, the key challenge is the water zone. Plan your horizontal moves so that when you move UP you can land fully on a log or turtle. Even a small misalignment may cost a life, so careful lateral adjustments are crucial.",\\
    "Minimize Unnecessary Moves: Every lateral move that doesn’t contribute to safe alignment delays vertical progress. Use lateral moves only when necessary and focus on maintaining a steady upward progress when the safe gap in the next lane is clear.",\\
    "Monitor the Environment: Keep an eye on the positions of both the cars and the water platforms. As you get closer to the target (LadyFrog), adjust your position gradually rather than making large leaps, ensuring a smooth transition into the goal area.",\\
    "Practice Patience: In many instances, waiting (using NOOP) for the right moment before committing to an upward move can be beneficial. Let the obstacles move into a more favorable configuration instead of forcing your progress."\\
  ]\\
\}
\end{gradientbox}

\subsection{Samples from round 2 of o3-mini-2025-01-31 with high reasoning, 0 past steps, rewards shown, with feedback from round 1}
\label{pdf:o3-mini_high_past_all_rewards_show_round_2}
\includegraphics[width=\textwidth, trim=0 100 0 0, clip]{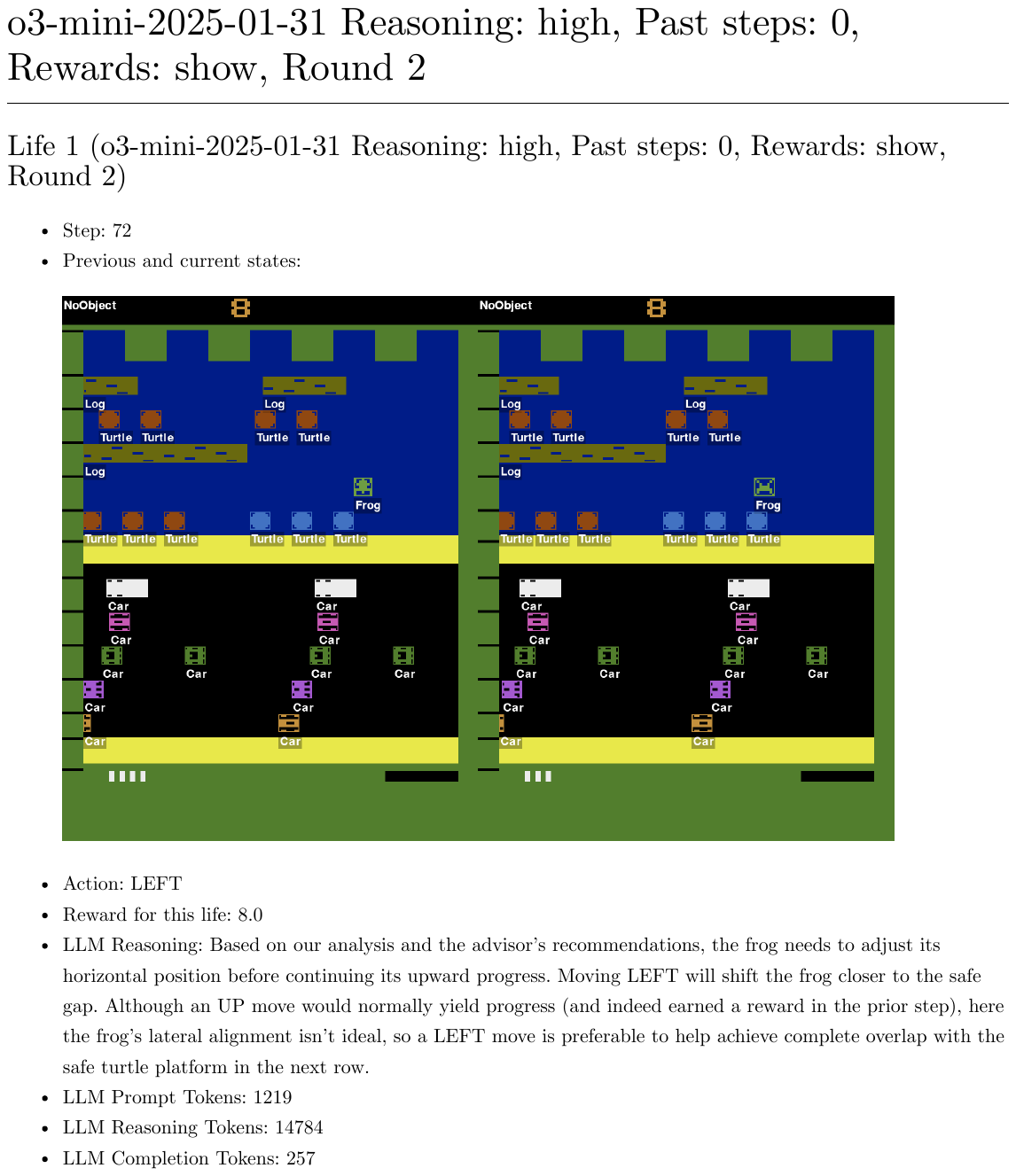}
\clearpage


\section*{Supplementary material}

## 5.8. Samples from round 2 of o3-mini-2025-01-31 with high reasoning, 0 past steps, rewards shown, with feedback from round 1

# o3-mini-2025-01-31 Reasoning: high, Past steps: 0, Rewards: show, Round 2

## Life 1 (o3-mini-2025-01-31 Reasoning: high, Past steps: 0, Rewards: show, Round 2)

- Step: 72
- Previous and current states:

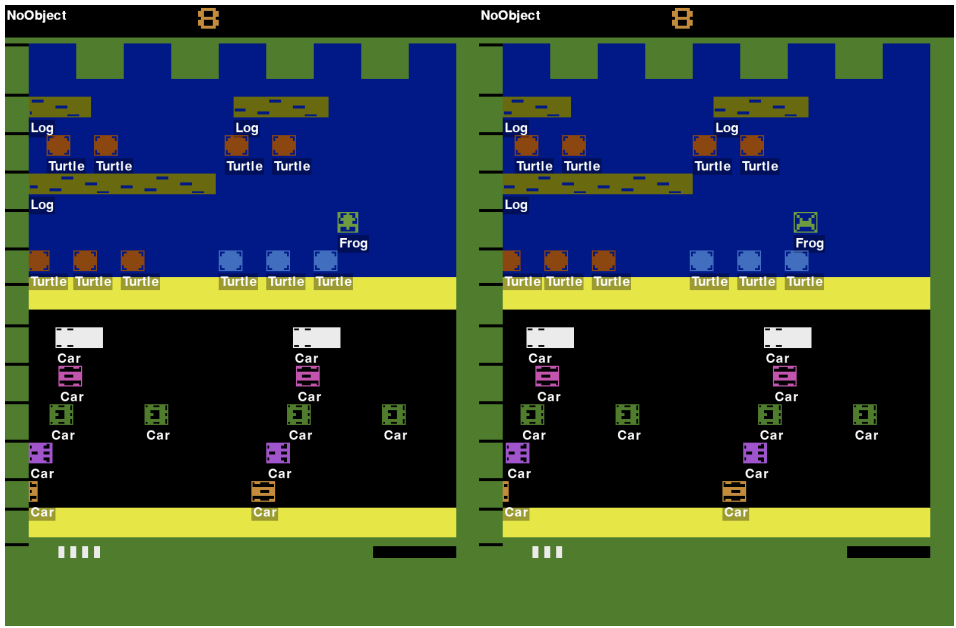

- Action: LEFT
- Reward for this life: 8.0
- LLM Reasoning: Based on our analysis and the advisor's recommendations, the frog needs to adjust its horizontal position before continuing its upward progress. Moving LEFT will shift the frog closer to the safe gap. Although an UP move would normally yield progress (and indeed earned a reward in the prior step), here the frog's lateral alignment isn't ideal, so a LEFT move is preferable to help achieve complete overlap with the safe turtle platform in the next row.
- LLM Prompt Tokens: 1219
- LLM Reasoning Tokens: 14784
- LLM Completion Tokens: 257

# Life 2 (o3-mini-2025-01-31 Reasoning: high, Past steps: 0, Rewards: show, Round 2)

- Step: 88
- Previous and current states:

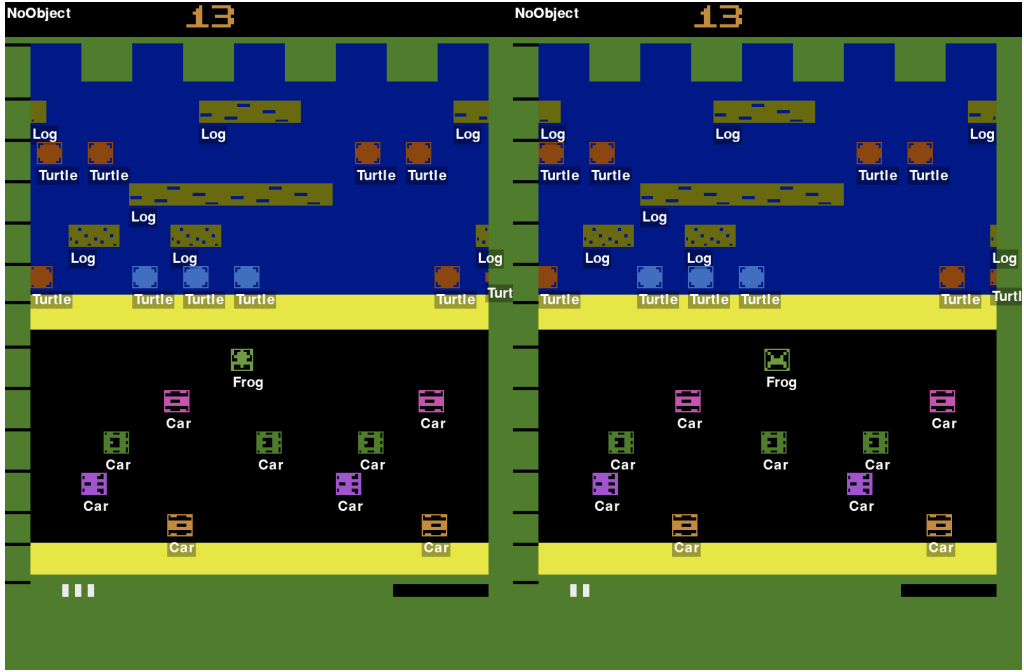

- Action: RIGHT
- Reward for this life: 5.0
- LLM Reasoning: Although moving UP has yielded a positive reward so far, advancing now upward without proper horizontal alignment risks landing partially off the safe platform. The advice is to make a lateral adjustment before continuing upward when the frog isn't fully aligned with a safe gap. Since the target turtle's safe region ideally requires the frog's x-coordinate to be 73 or 74, a RIGHT move will move the frog closer to that range. LEFT or NOOP would not improve the alignment and DOWN would be against progress.
- LLM Prompt Tokens: 1219
- LLM Reasoning Tokens: 8960
- LLM Completion Tokens: 273

# Life 3 (o3-mini-2025-01-31 Reasoning: high, Past steps: 0, Rewards: show, Round 2)

- Step: 102
- Previous and current states:

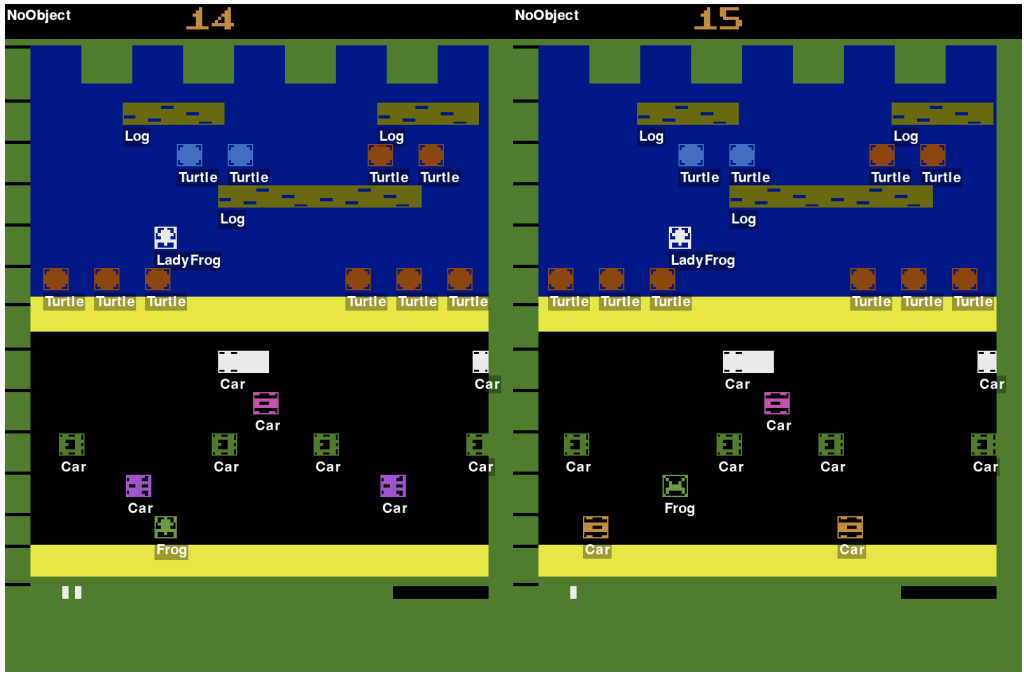

- Action: UP
- Reward for this life: 2.0
- LLM Reasoning: Given that the last upward move was rewarded and the frog's horizontal position is already well aligned with the safe gap in the upcoming lane (and nearly with LadyFrog), moving UP again maximizes vertical progress without introducing lateral risks. Lateral moves aren't necessary at this moment because the frog is clear of obstacles in the gap, and delaying with NOOP or moving LEFT/RIGHT could misalign it. Thus, the best action is to continue moving upward.
- LLM Prompt Tokens: 1191
- LLM Reasoning Tokens: 5952
- LLM Completion Tokens: 245

# Life 4 (o3-mini-2025-01-31 Reasoning: high, Past steps: 0, Rewards: show, Round 2)

- Step: 114
- Previous and current states:

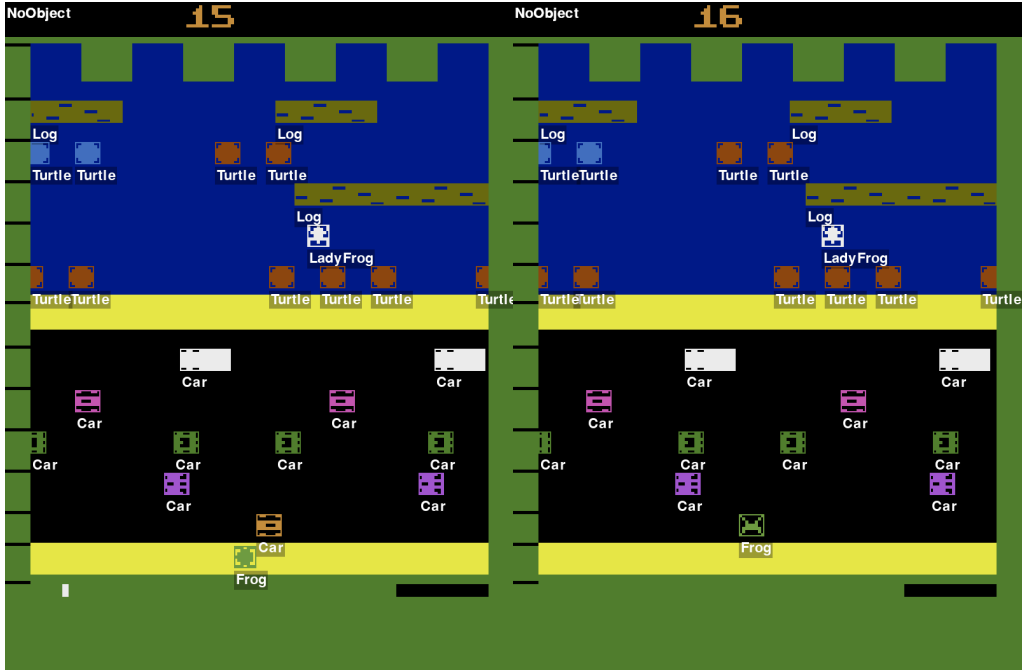

- Action: UP
- Reward for this life: 1.0
- LLM Reasoning: According to the advice, upward moves are the main driver of progress when the frog is positioned in a safe gap. The frog's current horizontal coordinate (72) keeps it clear of the car in the next lane (car at (80,161)), ensuring no dangerous overlap when moving up. Moving right now might risk lateral collision when the frog advances into a lane with a moving car, and further left or downward delays progress. Therefore, the best action is to move UP, maintaining safe progress and planning for later horizontal adjustments to better align with LadyFrog.
- LLM Prompt Tokens: 1221
- LLM Reasoning Tokens: 7424
- LLM Completion Tokens: 291

# Life 5 (o3-mini-2025-01-31 Reasoning: high, Past steps: 0, Rewards: show, Round 2)

- Step: 133
- Previous and current states:

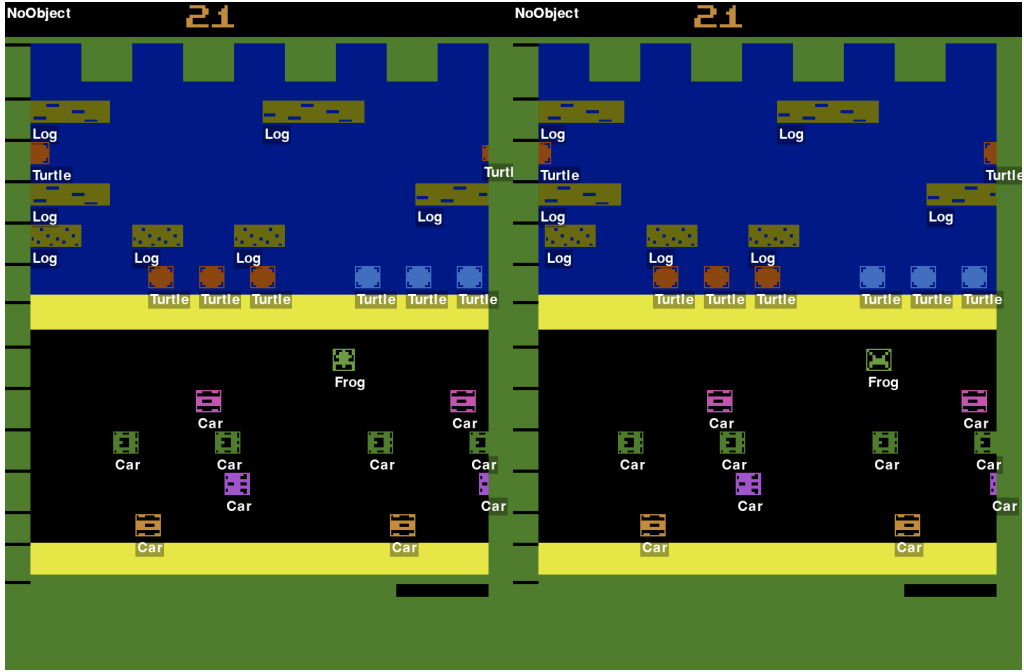

- Action: RIGHT
- Reward for this life: 5.0
- LLM Reasoning: Moving upward immediately (UP) would continue vertical progress but risks landing in a water lane without adequate support because the frog's current horizontal alignment does not overlap either safe turtle cluster. Similarly, a LEFT move would shift the frog further away from a safe platform since the left cluster is far to the left, while down would regress progress. Based on pre-analysis advice emphasizing proper horizontal alignment before advancing upward, a RIGHT move will reposition the frog closer to the right safe cluster (e.g., the turtle at (111,83)) where its hitbox (after shifting from 103 to around 110) will overlap significantly. This proactive lateral adjustment maximizes the chance of landing completely on a safe platform in the upcoming water lane.
- LLM Prompt Tokens: 1204
- LLM Reasoning Tokens: 7232
- LLM Completion Tokens: 331



\end{document}